\documentclass[letterpaper]{article}
\usepackage[preprint]{1}
\usepackage[hyphens]{url}
\usepackage{graphicx}
\usepackage{natbib}
\usepackage{caption}
\usepackage{amsmath}
\usepackage{amssymb}
\usepackage{subcaption}
\usepackage{booktabs}
\usepackage{multirow}
\usepackage{colortbl}
\usepackage{xcolor}
\usepackage{placeins}

\title{MA-DAR: Manifold-Aligned Dynamic Adaptive Routing for Continual Temporal Knowledge Graph Reasoning}

\author{
  Xiangjun Shi\textsuperscript{1},
  Chong Mu\textsuperscript{1},
  Jinchuan Zhang\textsuperscript{1,*}\thanks{Corresponding author.},
  Lizong Zhang\textsuperscript{1},
  Yuefeng He\textsuperscript{1},
  Shang Liu\textsuperscript{1}
}
\affiliations{
  \textsuperscript{1}School of Computer Science and Engineering, University of Electronic Science and Technology of China
}

\begin{document}
\maketitle

\begin{abstract}
Continual temporal knowledge graph (TKG) reasoning aims to continuously incorporate newly emerging facts while preserving previously acquired knowledge. Replay-based continual learning has achieved promising performance by revisiting historical representations. However, existing methods primarily focus on what to replay, while largely overlooking how replayed representations should be integrated with current ones. Such direct integration often gives rise to two critical forms of representation conflict: \textit{norm domination} and \textit{semantic blurring}, ultimately degrading continual reasoning performance.
To address these challenges, we propose MA-DAR (Manifold-Aligned Dynamic Adaptive Routing), a lightweight plug-and-play framework for replay representation fusion. MA-DAR first aligns replayed and current representations onto a shared manifold to alleviate distribution discrepancies. It then employs a dynamic gating mechanism to learn dimension-wise fusion weights, adaptively determining the contribution of replayed and current representations to the fused representation. Furthermore, a polarization regularizer encourages more decisive routing behaviors by discouraging ambiguous gating decisions, resulting in more stable and effective knowledge integration.
Extensive experiments on four public continual TKG benchmarks demonstrate that MA-DAR consistently improves the performance of representative TKG encoders while remaining effective under different replay settings. Comprehensive ablation studies and visualization analyses further verify the effectiveness of manifold alignment and dynamic adaptive routing in mitigating representation conflicts and improving continual reasoning.
\end{abstract}

\section{Introduction}
Temporal Knowledge Graphs (TKGs) have emerged as highly effective structures for capturing the evolutionary dynamics of facts across time, underpinning critical downstream applications such as event forecasting and intelligent risk analysis \citep{huang2024confidence, chen2024unified, zheng2020privacy}. A TKG represents facts as sequences of quadruples, comprising subject, relation, object, and timestamp. A primary challenge in this domain is extrapolation reasoning: inferring missing facts at unseen future timestamps based on historical topological patterns \citep{lacroix2020tensor, goel2020diachronic, li2021temporal, jin2020recurrent}.

Real-world TKG data typically arrives in continuous, streaming snapshots. Sequential training on such streaming graphs natively induces catastrophic forgetting \citep{kirkpatrick2017overcoming}, where the model abruptly loses previously acquired historical dependencies. To address this, considerable efforts have been focused on updating graph encoders to capture richer and more expressive structural representations. For instance, RE-GCN models local structural evolution \citep{li2021temporal}, TiRGN \citep{li2022tirgn} incorporates local-global historical patterns, and LogCL \citep{meng2023logcl} introduces contrastive learning to alleviate data sparsity. Meanwhile, experience replay methods attempt to preserve historical knowledge by revisiting past information during incremental updates. For instance, ER \citep{rolnick2019experience} mitigates forgetting by storing and rehearsing a raw subset of historical facts, while DGAR \citep{zhang2025generative} explores generative replay by producing pseudo-historical distributions through a diffusion model.

\begin{figure}[t]
  \centering
  \includegraphics[width=\linewidth]{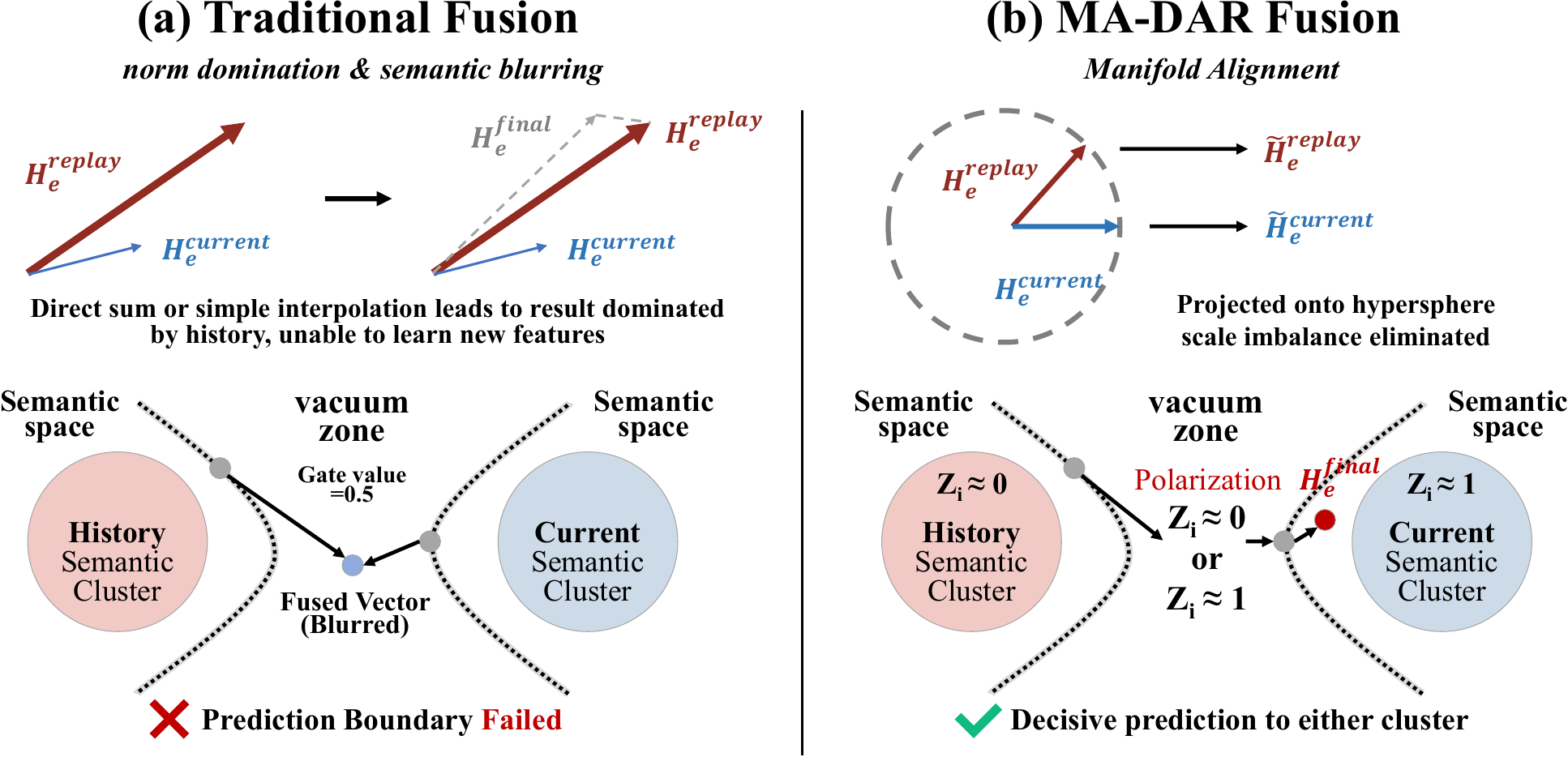} 
  \caption{Comparison of feature fusion mechanisms in TKG reasoning.}
  \label{fig:intro}
\vspace{-5mm}
\end{figure}

However, existing replay-based methods mainly focus on what to replay, while paying less attention to how replayed representations should be integrated with current representations. Most baselines rely on global scalar weights or linear concatenation to merge the historical replay representation with the current topological update \citep{wu2021tie, cui2023lifelong}. This coarse-grained fusion strategy ignores the semantic complexity of high-dimensional embeddings, leading to two critical representation conflicts: \textit{norm domination} and \textit{semantic blurring}. Specifically, generative replay methods such as DGAR \citep{zhang2025generative} may introduce scale discrepancies during the reconstruction of historical embeddings, causing the replayed representations to dominate the fusion process. As depicted in Figure \ref{fig:intro}(a), such magnitude imbalance weakens the contribution of current structural updates during representation integration. Meanwhile, without dimension-aware constraints, conventional gating mechanisms tend to produce ambiguous weights around 0.5, resulting in insufficient separation between historical information and current dynamics and ultimately blurring the synthesized semantic space.

In this paper, we propose the Manifold-Aligned Dynamic Adaptive Routing (MA-DAR). Diverging from standard end-to-end architectures, MA-DAR acts as a generic, plug-and-play routing plugin that decouples replay representation fusion from current-step structural encoding. As depicted in Figure \ref{fig:intro}(b), MA-DAR establishes dual data interfaces to ingest both the current temporal state from an arbitrary base encoder (e.g., RE-GCN, TiRGN, LogCL) and the long-term historical representations provided by an independent replay model, executing localized feature rectification within its own module. To overcome fusion bottlenecks, the MA-DAR plugin executes a compact sequence of operations: Manifold Alignment first normalizes historical and current representations to map them without norm domination; Dynamic Gating then aligns semantics by calculating distinct weights for individual dimensions; and finally, a Polarization Penalty regularizes the fusion process to force explicit, binary-like retention decisions. 

Our contributions are summarized as follows:
\begin{itemize}
\item We propose the MA-DAR framework, a lightweight and plug-and-play replay representation fusion framework that identifies and alleviates two critical representation conflicts, namely \textit{norm domination} and \textit{semantic blurring}, in continual TKG reasoning.
\item We design a multi-stage routing protocol inside the plugin, blending Manifold Alignment, Dynamic Gating, and a differentiable Polarization Penalty to eliminate norm discrepancy and achieve sharp, binary-like feature decoupling.
\item We conduct extensive evaluations across multiple continual TKG benchmarks, demonstrating that MA-DAR achieves state-of-the-art performance while remaining effective under different replay settings and TKG encoders.
\end{itemize}

\section{Related Work}
\subsection{Specialized Architectural Optimization}
Traditional and early temporal knowledge graph reasoning approaches captured evolutionary dynamics via static time embeddings \citep{bordes2013translating, leblay2018deriving, dasgupta2018hyte} or autoregressive sequence layers \citep{jin2020recurrent}. Subsequently, message-passing schemes via structural GNNs became prevalent, with RE-GCN capturing local evolution \citep{li2021temporal} and TiRGN modeling local-global periodic patterns \citep{li2022tirgn}. More recently, deep-tuning paradigms have introduced advanced multi-granularity and contrastive regularizers to alleviate topological sparsity \citep{xu2023temporal, meng2023logcl, chen2024local, wang2024federal}, while others leverage dual correspondence or latent relational associations to mine periodic patterns \citep{liang2023learn, zhang2023learning}. 
To capture unobserved topologies, contemporary architectures incorporate multi-graph convolutions or historically relevant event structures to dynamically prune dependency paths \citep{zhang2024learning, zhang2025historically}. Despite their expressive structural encoding capabilities, these models generally lack explicit memory retention mechanisms; continuous streaming updates may gradually overwrite previously captured patterns, leading to catastrophic forgetting.

\subsection{Regularization and Distillation}
Regularization-based paradigms penalize parameter or representation deviation to preserve historical weight locations \citep{kirkpatrick2017overcoming, zenke2017continual, aljundi2018memory}. Early alignment methods enforced consistency via embedding coordinate tracking \citep{wang2019sentence}. In graph domains, these constraints manifest as temporal consistency metrics \citep{wu2021tie}, $L_2$-norm penalties over expanding structures \citep{cui2023lifelong}, or specialized debiasing paths to counter spurious forgetting \citep{zhu2025debiasedkge}. 

Beyond weight penalties, knowledge distillation and evolutionary tokens act as structural regularizers, inheriting the foundational principle of learning without forgetting \citep{li2017learning}. Modern variants align incremental graphs via incremental distillation from offline teachers \citep{liu2024towards} or handle entity growth through scale-aware gradual evolution \citep{li2025sage}. To optimize parameter efficiency during streaming, task-driven tokens \citep{zhu2025ett} and selective incremental subgraph training \citep{jia2025sit} have been explored. Local-global structural distillation also serves to simultaneously constrain multi-granularity graph semantics \citep{shi2025continual}. Nevertheless, these continuous scalar penalties inherently force representations into a compromised intermediate state, failing to cleanly separate historical stability from current plasticity.

\subsection{Experience Replay}
Experience replay mitigates forgetting by interleaving stored or generated historical distributions with current streaming updates \citep{lopez2017gradient, rolnick2019experience, daruna2021continual}. To accommodate continuous relational drift, advanced frameworks integrate external guidance from large language models \citep{wang2024large} or utilize dynamic state tracking to update representations \citep{zhang2023streame}. Concurrently, generative replay based on conditional score-matching and denoising diffusion probabilistic models has emerged to synthesize high-fidelity historical contexts resilient to long-term drift \citep{ho2020denoising, austin2021structured, long2024fact, cai2024predicting, zhang2025generative, chen2026lifelong}. However, a critical bottleneck remains: many existing replay-based TKG methods rely on coarse global scalar weights or vector concatenation to fuse replayed distributions with current states. This ignores high-dimensional scale disparities, inducing severe \textit{norm domination} and \textit{semantic blurring}, which motivates the need for a more adaptive replay representation fusion mechanism.

\section{Problem Formulation}

A Temporal Knowledge Graph $\mathcal{G}$ is formalized as a sequence of snapshots $\mathcal{G} = \{G_1, G_2, \dots, G_{|\mathcal{T}|}\}$. Each snapshot $G_t$ contains a set of quadruples $(s, r, o, t)$. In the streaming TKG reasoning setting, the model is tasked with answering queries $(s, r, ?, t)$ or $(?, r, o, t)$ given the historical graphs up to $t-1$. The model is continuously trained on sequentially arriving snapshots and evaluated on its ability to perform continual reasoning while preserving previously acquired knowledge.

\section{Methodology}
\begin{figure*}[t]
  \centering
  \includegraphics[width=1.0\textwidth]{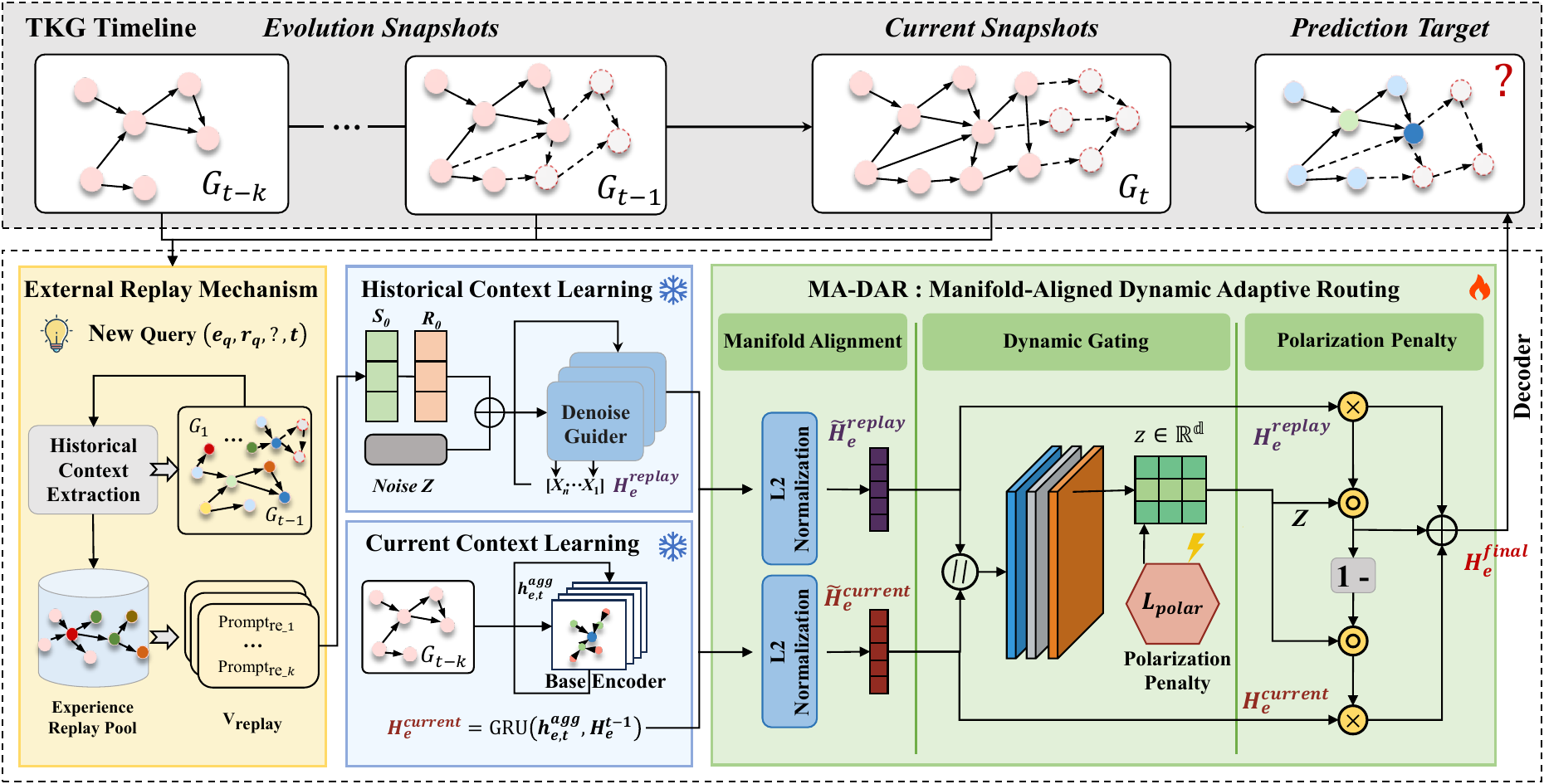} 
  \caption{Overall architecture of the MA-DAR plugin. Operating as a plug-and-play extension, MA-DAR establishes dual data interfaces to ingest external current temporal features and replayed historical representations. These heterogeneous context representations are sequentially aligned and routed within the plugin through Manifold Alignment, Dynamic Gating, and a Polarization Penalty.}
  \label{fig:framework}
  \vspace{-5mm}
  \end{figure*}

The overall architecture of the proposed Manifold-Aligned Dynamic Adaptive Routing (MA-DAR) framework is illustrated in Figure \ref{fig:framework}. Architecturally, MA-DAR is designed as a standalone modular plugin that establishes dual data interfaces to seamlessly couple with two independent external input streams: Current Encoding and Historical Replay Representation. Upon receiving these heterogeneous feature representations, the MA-DAR plugin routes them through its core components to systematically alleviate \textit{norm domination} and \textit{semantic blurring}.

\subsection{Current Encoding}
To capture the short-term evolutionary dynamics of the current TKG snapshot, MA-DAR interfaces with a base structural encoder $\mathcal{E}$. Let $\mathcal{G}_{t-m:t-1}$ denote the sequence of recent historical subgraphs. At each timestamp, the base encoder typically employs a multi-relational Graph Convolutional Network (GCN) to aggregate local neighborhood information. For a given entity $e$, its structural representation at layer $l$ is updated via message passing:
\begin{equation}
    h_{e}^{(l)} = \sigma \left( \sum_{r \in \mathcal{R}} \sum_{v \in \mathcal{N}_r(e)} W_r^{(l)} h_v^{(l-1)} \right)
\end{equation}
where $\mathcal{N}_r(e)$ is the set of neighbors connected by relation $r$, and $W_r^{(l)}$ is the relation-specific weight matrix. 

Subsequently, a Gated Recurrent Unit (GRU) evolves the entity state over the temporal window to capture dynamic trends, producing the current-step representation $H_{current}^{(t)}$:
\begin{equation}
    H_{current}^{(t)} = \text{GRU}(H_{GCN}^{(t)}, H_{current}^{(t-1)})
\end{equation}
The mathematical independence of MA-DAR from the internal mechanics of $\mathcal{E}$ empowers it to seamlessly interface with diverse representative TKG encoders, treating their final output identically as $H_{current}^{(t)}$.

\subsection{Historical Replay Representation}

Relying solely on recent topological updates leads to the catastrophic forgetting of long-term patterns. To provide a robust historical counterpart to $H_{current}^{(t)}$, MA-DAR receives historical replay representations $H_{replay}^{(t)}$ provided by external replay mechanisms. In our implementation, we instantiate the replay mechanism with the diffusion-based generative replay strategy adopted in DGAR as a representative source of historical representations.

In our diffusion-based instantiation, we follow DGAR to obtain the historical replay representation through a guided reverse-denoising process conditioned on the temporal query. The resulting representation $H_{\mathrm{replay}}^{(t)}$ is treated as an external input to MA-DAR. Importantly, MA-DAR is agnostic to the internal form of the replay mechanism and can also operate with directly sampled experience replay, as evaluated in RQ3.

\subsection{Components of MA-DAR}
Upon receiving $H_{\text{current}}^{(t)}$ and $H_{\text{replay}}^{(t)}$ via its dual entry interfaces, the MA-DAR plugin processes these representations through three sequentially cascaded alignment and gating steps.

\subsubsection{Manifold Alignment}
Vectors of disparate scales skew the fusion result, causing the magnitude-heavy representation to overshadow the other. To eliminate this \textit{norm domination} prior to integration, MA-DAR projects both the current and replay representations onto a uniform hypersphere manifold:
\begin{align}
    \tilde{H}_{current} &= \frac{H_{current}}{\|H_{current}\|_2 + \epsilon} \\
    \tilde{H}_{replay} &= \frac{H_{replay}}{\|H_{replay}\|_2 + \epsilon}
\end{align}
where $\epsilon$ is a small constant for numerical stability. This normalization removes norm-based scale advantages before the subsequent routing operation. The normalization is applied independently to each entity representation.

\subsubsection{Dynamic Gating}
To alleviate the \textit{semantic blurring} caused by ambiguous global scalar weights, MA-DAR computes a fine-grained, dimension-level gating vector $z \in [0, 1]^d$ using a shallow multilayer perceptron (MLP):
\begin{equation}
    z = \sigma(\text{MLP}(\tilde{H}_{replay} \oplus \tilde{H}_{current}))
\end{equation}
where $\oplus$ denotes concatenation and $\sigma$ is the sigmoid activation function. This vector evaluates information at the feature-wise level, allowing the model to adaptively route specific semantic dimensions from either the historical distribution or the current context. Finally, the adaptive semantic synthesis is performed using this gating vector:
\begin{equation}
    H_{final} = z \odot \tilde{H}_{replay} + (1-z) \odot \tilde{H}_{current}
\end{equation}
where $\odot$ represents the element-wise Hadamard product. 

\subsubsection{Polarization Penalty}
Without additional structural constraints, the routing vector $z$ calculated in the gating stage may remain in ambiguous intermediate states (e.g., values near 0.5), leading to less decisive retention behaviors. To force definitive choices and prevent features from blurring into a noisy superposition state, we introduce a strict Polarization Penalty $\mathcal{L}_{polar}$ during the optimization phase:
\begin{equation}
\mathcal{L}_{polar}
=
\frac{1}{|\mathcal{B}|d}
\sum_{e\in\mathcal{B}}
\sum_{j=1}^{d}
z_{e,j}(1-z_{e,j}),
\end{equation}
where $\mathcal{B}$ denotes the set of entities in the current mini-batch and $z_{e,j}$ is the gate value of entity $e$ at dimension $j$.
This regularizer penalizes ambiguous intermediate gate values and encourages each dimension toward more decisive values near $0$ or $1$, without imposing non-differentiable hard routing decisions.

\subsection{Joint Training Objective}
After obtaining the final polarized representation $H_{final}$, the decoder evaluates scores for candidate entities. We treat entity prediction as a multi-class classification task. For a streaming snapshot at task $t$, the task-specific training loss $\mathcal{L}_{task}$ is optimized via cross-entropy:
\begin{equation}
\mathcal{L}_{task}
=
-
\sum_{q\in\mathcal{D}_{train}^{(t)}}
\sum_{e\in\mathcal{E}}
y_{q,e}
\log p_{\theta}(e\mid q),
\label{eq:task_loss}
\end{equation}
where $q$ denotes an entity prediction query, $y_{q,e}$ is its one-hot ground-truth label, and $p_{\theta}(e\mid q)$ is the predicted probability of candidate entity $e$.

To counteract the historical information loss caused by overfitting to current temporal facts, we incorporate historical replay facts as an auxiliary replay objective. The loss associated with historical replay, $\mathcal{L}_{replay}$, is calculated identically to Equation \ref{eq:task_loss}, with the difference that it computes classification errors over historical facts sampled from $\mathcal{P}_{\text{replay}}$ instead of current snapshots. 

By incorporating the proposed dimension-level polarization regularizer while maintaining the original optimization objective of the backbone encoder, the joint training objective $\mathcal{L}$ for the entire framework is formalized as follows:
\begin{equation}
    \mathcal{L} = \mathcal{L}_{task} + \mu \mathcal{L}_{replay} + \lambda \mathcal{L}_{polar} + \alpha \mathcal{L}_{orig}
\end{equation}
where $\mu$, $\lambda$, and $\alpha$ control the contributions of historical replay, dimensional polarization, and the original optimization objective of the backbone encoder, respectively.

The additional computational overhead introduced by MA-DAR is negligible, as it only involves lightweight manifold alignment, dynamic gating, and polarization operations. Empirical profiling shows that the MA-DAR routing module accounts for less than 1\% of the total training time across both ER and diffusion-based replay settings. In the profiled diffusion-based configurations, it further accounts for only 0.18\%-0.22\% of the total inference time. Detailed efficiency analysis and hardware profiling are provided in the supplementary material.

\begin{table*}[t]
\centering
\caption{Performance (in percentage) comparisons on ICEWS14s, ICEWS18, GDELT, and ICEWS05-15 datasets under the time-aware filter setting. Bold values indicate the best performance in each base encoder group.}
\label{tab:main_results}
\resizebox{\textwidth}{!}{
\begin{tabular}{c|ccc|ccc|ccc|ccc}
\toprule
\multirow{2}{*}{\textbf{Method}} & \multicolumn{3}{c|}{\textbf{ICEWS14s}} & \multicolumn{3}{c|}{\textbf{ICEWS18}} & \multicolumn{3}{c|}{\textbf{GDELT}} & \multicolumn{3}{c}{\textbf{ICEWS05-15}} \\
\cmidrule(lr){2-4} \cmidrule(lr){5-7} \cmidrule(lr){8-10} \cmidrule(lr){11-13}
& \textbf{MRR} & \textbf{H@1} & \textbf{H@10} & \textbf{MRR} & \textbf{H@1} & \textbf{H@10} & \textbf{MRR} & \textbf{H@1} & \textbf{H@10} & \textbf{MRR} & \textbf{H@1} & \textbf{H@10} \\
\midrule
FT & 37.46 & 26.95 & 58.20 & 25.35 & 15.97 & 44.36 & 15.60 & 8.73 & 29.05 & 41.88 & 30.55 & 63.82 \\
ER & 42.14 & 31.03 & 63.80 & 27.20 & 16.88 & 48.19 & 16.21 & 8.97 & 30.42 & 45.55 & 33.34 & 69.07 \\
TIE & 41.07 & 30.28 & 62.39 & 28.73 & 18.40 & 49.60 & 16.40 & 8.94 & 30.98 & 42.56 & 30.90 & 64.67 \\
LKGE & 37.51 & 27.13 & 58.51 & 25.56 & 16.12 & 44.70 & 15.52 & 8.69 & 28.90 & 42.46 & 30.99 & 64.51 \\
IncDE & 36.57 & 26.20 & 56.95 & 25.52 & 16.07 & 44.74 & 15.49 & 8.64 & 28.86 & 40.56 & 29.34 & 62.17 \\
L$^2$TKG & 47.40 & 35.36 & 71.05 & 33.36 & 22.15 & 55.04 & 20.53 & 12.89 & 35.83 & 57.43 & 41.86 & 80.69 \\
RPC & 44.55 & 34.87 & 65.08 & 34.91 & 24.34 & 55.89 & 22.41 & 14.42 & 38.33 & 51.14 & 39.47 & 71.75 \\
HisRES & 50.48 & 39.57 & 71.09 & 37.69 & 26.46 & 59.70 & 26.58 & 16.90 & 46.31 & 59.07 & 48.62 & 78.48 \\
DGAR & 50.12 & 39.36 & 70.48 & 33.00 & 21.74 & 55.63 & 28.30 & 17.38 & 51.39 & 54.33 & 43.11 & 75.13 \\
\midrule
RE-GCN & 41.50 & 30.86 & 62.47 & 30.55 & 20.00 & 51.46 & 19.31 & 11.99 & 33.59 & 46.41 & 35.17 & 67.64 \\
RE-GCN+MA-DAR & \textbf{52.51} & \textbf{41.74} & \textbf{72.50} & \textbf{36.55} & \textbf{24.51} & \textbf{61.05} & \textbf{28.71} & \textbf{17.44} & \textbf{52.67} & \textbf{55.52} & \textbf{44.55} & \textbf{76.34} \\
\rowcolor{gray!15} \textit{$\Delta$ Improve} & \textit{+26.53\%} & \textit{+35.26\%} & \textit{+16.06\%} & \textit{+19.64\%} & \textit{+22.55\%} & \textit{+18.64\%} & \textit{+48.68\%} & \textit{+45.45\%} & \textit{+56.80\%} & \textit{+19.63\%} & \textit{+26.67\%} & \textit{+12.86\%} \\
\midrule
TiRGN & 44.04 & 33.83 & 63.84 & 33.66 & 23.19 & 54.22 & 21.67 & 13.63 & 37.60 & 50.04 & 39.25 & 70.71 \\
TiRGN+MA-DAR & \textbf{52.40} & \textbf{42.20} & \textbf{71.30} & \textbf{35.56} & \textbf{23.63} & \textbf{59.75} & \textbf{29.04} & \textbf{17.56} & \textbf{52.48} & \textbf{56.54} & \textbf{44.92} & \textbf{75.51} \\
\rowcolor{gray!15} \textit{$\Delta$ Improve} & \textit{+18.98\%} & \textit{+24.74\%} & \textit{+11.69\%} & \textit{+5.64\%} & \textit{+1.90\%} & \textit{+10.20\%} & \textit{+34.01\%} & \textit{+28.83\%} & \textit{+39.57\%} & \textit{+12.99\%} & \textit{+14.45\%} & \textit{+6.79\%} \\
\midrule
LogCL & 48.87 & 37.76 & 70.26 & 35.67 & 24.53 & 57.74 & 23.75 & 14.64 & 42.33 & 57.04 & 46.07 & 77.87 \\
LogCL+MA-DAR & \textbf{66.20} & \textbf{55.40} & \textbf{85.50} & \textbf{46.12} & \textbf{31.85} & \textbf{74.72} & \textbf{29.56} & \textbf{17.94} & \textbf{54.32} & \textbf{64.89} & \textbf{53.97} & \textbf{84.96} \\
\rowcolor{gray!15} \textit{$\Delta$ Improve} & \textit{+35.46\%} & \textit{+46.72\%} & \textit{+21.69\%} & \textit{+29.30\%} & \textit{+29.84\%} & \textit{+29.41\%} & \textit{+24.46\%} & \textit{+22.54\%} & \textit{+28.33\%} & \textit{+13.76\%} & \textit{+17.15\%} & \textit{+9.10\%} \\
\bottomrule
\end{tabular}
}
\vspace{-5mm}
\end{table*}

\section{Experiments}
To systematically evaluate the proposed framework, our experiments are designed to answer the following five Research Questions (RQs):

\begin{itemize}
\item \textbf{RQ1 (Superiority):} Does MA-DAR consistently enhance representative base TKG encoders in streaming scenarios?

\item \textbf{RQ2 (Ablation):} Do the core components of Manifold Alignment, Dynamic Gating, and Polarization Penalty contribute synergistically to performance improvements?

\item \textbf{RQ3 (Replay Compatibility):} Can MA-DAR effectively integrate historical representations generated by different replay mechanisms?

\item \textbf{RQ4 (Norm Analysis):} How does Manifold Alignment alleviate scale imbalance and resolve \textit{norm domination}?

\item \textbf{RQ5 (Gating Analysis):} How does the Polarization Penalty mitigate \textit{semantic blurring} and encourage decisive dimension-wise routing?
\end{itemize}

\subsection{Experimental Setup}

\subsubsection{Datasets}
We evaluate the proposed MA-DAR framework on four widely used Temporal Knowledge Graph benchmark datasets: ICEWS14s, ICEWS18, ICEWS05-15, and GDELT. The detailed statistics of these temporal knowledge graph benchmarks, including their chronological snapshots and temporal granularities, are provided in the supplementary material.

\subsubsection{Baselines}
We compare MA-DAR against representative state-of-the-art baselines from different paradigms: (1) continual knowledge graph reasoning methods, including FT (Fine-Tuning), ER \citep{rolnick2019experience}, TIE \citep{wu2021tie}, LKGE \citep{cui2023lifelong}, IncDE \citep{liu2024towards}, and DGAR \citep{zhang2025generative}; and (2) representative TKG reasoning methods with specialized modeling strategies, including RPC \citep{liang2023learn}, L$^2$TKG \citep{zhang2023learning}, and HisRES \citep{zhang2025historically}. To demonstrate the plug-and-play capability of MA-DAR, we integrate it with diverse representative base encoders, including RE-GCN \citep{li2021temporal}, TiRGN \citep{li2022tirgn}, and LogCL \citep{meng2023logcl}.

\subsection{Implementation Details}

Following common settings in continual TKG reasoning, the embedding dimension $d$ is set to 200, the learning rate is set to 0.001, and optimization is performed using Adam. The historical replay sample size $k$ is configured to 35, 25, 40, and 32 for ICEWS14s, ICEWS18, ICEWS05-15, and GDELT, respectively, with $\mu=1$.

For the internal routing operations within MA-DAR, the stability constant $\epsilon$ in Manifold Alignment is fixed at $1\times10^{-5}$. The Dynamic Gating network is implemented as a shallow MLP with sigmoid activation. Its hidden dimension, number of layers, and dropout rate are selected according to the adopted backbone encoder. The polarization penalty coefficient is set to $\lambda=0.2$ for RE-GCN and $\lambda=0.5$ for TiRGN and LogCL. Detailed backbone-specific configurations are provided in the supplementary material. To ensure fair evaluation, the auxiliary loss coefficient $\alpha$ follows the default structural objectives of different base encoders: $\alpha=0.1$ for RE-GCN's static constraints, $\alpha=1.0$ for TiRGN's evolution loss, and $\alpha=0.5$ for LogCL's graph contrastive objective. Detailed hyperparameter analysis is provided in the supplementary material. To assess robustness to random initialization, we repeat the representative RE-GCN+MA-DAR configuration three times with different random seeds on all four datasets. The mean results are reported in the main paper, while the corresponding mean $\pm$ standard deviation results are provided in the supplementary material.

Additionally, the replay mechanism employed in the main experiments is the diffusion‑based generative replay. Under this setting, MA-DAR can be regarded as diffusion + RE-GCN.

\subsection{Main Results (RQ1)}
The continual reasoning performance across four benchmark datasets under the strict time-aware filter setting is presented in Table \ref{tab:main_results}. The upper section of the table reports the results of continual learning baselines and representative TKG reasoning methods with specialized modeling strategies, while the lower section presents the performance of representative base encoders equipped with our MA-DAR plugin to demonstrate the improvements brought by MA-DAR.

\begin{table*}[t]
\centering
\caption{Ablation study of the proposed MA-DAR on three datasets across representative base encoders. ``w/o'' denotes the removal of a specific component.}
\label{tab:ablation_all}
\renewcommand{\arraystretch}{0.8}
\resizebox{\textwidth}{!}{
\begin{tabular}{l|ccc|ccc|ccc}
\toprule
\multirow{2}{*}{\textbf{Method}} & \multicolumn{3}{c|}{\textbf{ICEWS14s}} & \multicolumn{3}{c|}{\textbf{ICEWS18}} & \multicolumn{3}{c}{\textbf{ICEWS05-15}} \\
\cmidrule(lr){2-4} \cmidrule(lr){5-7} \cmidrule(lr){8-10}
& \textbf{MRR} & \textbf{H@1} & \textbf{H@10} & \textbf{MRR} & \textbf{H@1} & \textbf{H@10} & \textbf{MRR} & \textbf{H@1} & \textbf{H@10} \\
\midrule
\multicolumn{10}{c}{\textit{Base Encoder: RE-GCN}} \\
\midrule
\quad \textit{w/o} MA (Manifold Alignment)   & 51.92 & 40.95 & 72.20 & 35.36 & 23.53 & 59.18 & 54.01 & 43.82 & 75.83 \\
\quad \textit{w/o} DG (Dynamic Gating)     & 51.44 & 40.61 & 71.20 & 34.91 & 23.16 & 58.62 & 54.68 & 43.27 & 75.49 \\
\quad \textit{w/o} PP (Polarization Penalty)& 52.28 & 41.34 & 72.36 & 36.21 & 24.17 & 60.68 & 55.13 & 44.17 & 76.04 \\
\rowcolor{gray!15} \textbf{+ MA-DAR (Full)} & \textbf{52.51} & \textbf{41.74} & \textbf{72.50} & \textbf{36.55} & \textbf{24.51} & \textbf{61.05} & \textbf{55.52} & \textbf{44.55} & \textbf{76.34} \\
\midrule
\multicolumn{10}{c}{\textit{Base Encoder: TiRGN}} \\
\midrule
\quad \textit{w/o} MA (Manifold Alignment)   & 51.75 & 41.36 & 70.98 & 34.42 & 22.68 & 57.85 & 55.08 & 44.15 & 74.98 \\
\quad \textit{w/o} DG (Dynamic Gating)     & 51.28 & 41.05 & 70.05 & 33.95 & 22.31 & 57.34 & 55.62 & 43.65 & 74.65 \\
\quad \textit{w/o} PP (Polarization Penalty)& 52.12 & 41.83 & 71.15 & 35.21 & 23.32 & 59.35 & 56.12 & 44.52 & 75.18 \\
\rowcolor{gray!15} \textbf{+ MA-DAR (Full)} & \textbf{52.40} & \textbf{42.20} & \textbf{71.30} & \textbf{35.56} & \textbf{23.63} & \textbf{59.75} & \textbf{56.54} & \textbf{44.92} & \textbf{75.51} \\
\midrule
\multicolumn{10}{c}{\textit{Base Encoder: LogCL}} \\
\midrule
\quad \textit{w/o} MA (Manifold Alignment)   & 58.15 & 46.22 & 78.30 & 40.12 & 27.55 & 66.85 & 59.15 & 48.82 & 79.25 \\
\quad \textit{w/o} DG (Dynamic Gating)     & 56.40 & 44.85 & 76.95 & 38.85 & 26.20 & 65.10 & 58.05 & 47.95 & 78.45 \\
\quad \textit{w/o} PP (Polarization Penalty)& 61.25 & 49.95 & 81.45 & 42.65 & 29.35 & 70.25 & 61.88 & 50.75 & 81.65 \\
\rowcolor{gray!15} \textbf{+ MA-DAR (Full)} & \textbf{66.20} & \textbf{55.40} & \textbf{85.50} & \textbf{46.12} & \textbf{31.85} & \textbf{74.72} & \textbf{64.89} & \textbf{53.97} & \textbf{84.96} \\
\bottomrule
\end{tabular}
}
\vspace{-5mm}
\end{table*}

Based on the quantitative results, several observations can be established. First, equipping diverse base encoders with MA-DAR consistently improves performance across different datasets. Regardless of whether the underlying model relies on local structural evolution modeling (RE-GCN), local-global periodic pattern modeling (TiRGN), or structural contrastive learning (LogCL), the consistent improvements from each \textit{Baseline} to \textit{Baseline}+MA-DAR counterpart in Table \ref{tab:main_results} demonstrate the general applicability of MA-DAR as a plug-and-play enhancement module. For instance, integrating MA-DAR with the basic RE-GCN model achieves a relative MRR improvement of \textbf{26.53\%} on ICEWS14s and a relative H@10 improvement of \textbf{56.80\%} on the high-frequency GDELT dataset under the same filter evaluation protocol.

Second, combining LogCL with MA-DAR achieves the strongest overall performance among all evaluated methods. Compared with competitive baselines such as HisRES and DGAR, LogCL+MA-DAR achieves an absolute MRR of \textbf{66.20\%} on ICEWS14s, outperforming HisRES by an absolute margin of \textbf{15.72\%}. It also obtains relative MRR improvements of \textbf{22.37\%} on ICEWS18 and \textbf{9.85\%} on ICEWS05-15 over HisRES. These results demonstrate the effectiveness of fine-grained dimension-level routing in MA-DAR for integrating historical and current representations.
\FloatBarrier

\subsection{Ablation Study (RQ2)}

To evaluate the contribution of the core components within the plugin, we examine three variants of MA-DAR across all representative encoders: (1) w/o MA (Manifold Alignment), which skips hypersphere projection and directly feeds raw embeddings into the routing module; (2) w/o DG (Dynamic Gating), which disables the dimension-level MLP router and replaces it with simple scalar averaging ($H_{final} = 0.5 \tilde{H}_{replay} + 0.5 \tilde{H}_{current}$); and (3) w/o PP (Polarization Penalty), which optimizes without the $\mathcal{L}_{polar}$ regularizer.

Quantitative results in Table \ref{tab:ablation_all} show that removing any individual component leads to performance degradation across all backbones, demonstrating the contribution of each component to the overall effectiveness of MA-DAR. Specifically, the w/o DG variant suffers the largest performance degradation (e.g., LogCL's MRR decreases by 9.80\% on ICEWS14s), indicating that scalar averaging is insufficient for effective representation integration. Meanwhile, the degradation of the w/o MA variant demonstrates that removing hypersphere alignment makes the fusion process more sensitive to scale discrepancies between historical and current representations. Finally, the performance decrease of the w/o PP variant indicates that removing polarization regularization weakens the ability of the gating module to make decisive dimension-wise routing decisions. Overall, these results demonstrate that Manifold Alignment, Dynamic Gating, and Polarization Penalty provide complementary benefits, and their combination enables MA-DAR to effectively integrate historical and current representations, answering RQ2.

\subsection{Replay Compatibility Analysis (RQ3)}
To investigate whether MA-DAR can effectively utilize historical representations from different replay mechanisms, we further evaluate its compatibility with both real experience replay (ER) and generative replay settings. Specifically, we combine MA-DAR with LogCL under different historical representation sources and compare the resulting performance.

\begin{table}[t]
\centering
\caption{Replay compatibility analysis of MA-DAR on ICEWS14s.}
\label{tab:replay_compatibility}
\begin{tabular}{c|c|c}
\midrule
Method & Replay & MRR \\
\midrule
LogCL & None & 48.87 \\
Diffusion + LogCL & Gen. & 51.58 \\
ER + LogCL + MA-DAR & Real & 51.70 \\
Diffusion + LogCL + MA-DAR & Gen. & 66.20 \\
\midrule
\end{tabular}
\vspace{-6mm}
\end{table}

The results are summarized in Table \ref{tab:replay_compatibility}. Under the real experience replay setting, MA-DAR improves LogCL from 48.87\% to 51.70\% MRR, demonstrating that the proposed alignment and routing mechanism remains effective when historical representations are obtained through direct sampling.
Furthermore, under the generative replay setting, integrating MA-DAR with LogCL further improves the MRR from 51.58\% to 66.20\%. The consistent improvements under different replay sources indicate that MA-DAR focuses on resolving representation integration conflicts rather than relying on a specific replay generation mechanism.

\subsection{Norm Domination Analysis (RQ4)}

\begin{figure}[t]
\centering
\includegraphics[width=\linewidth]{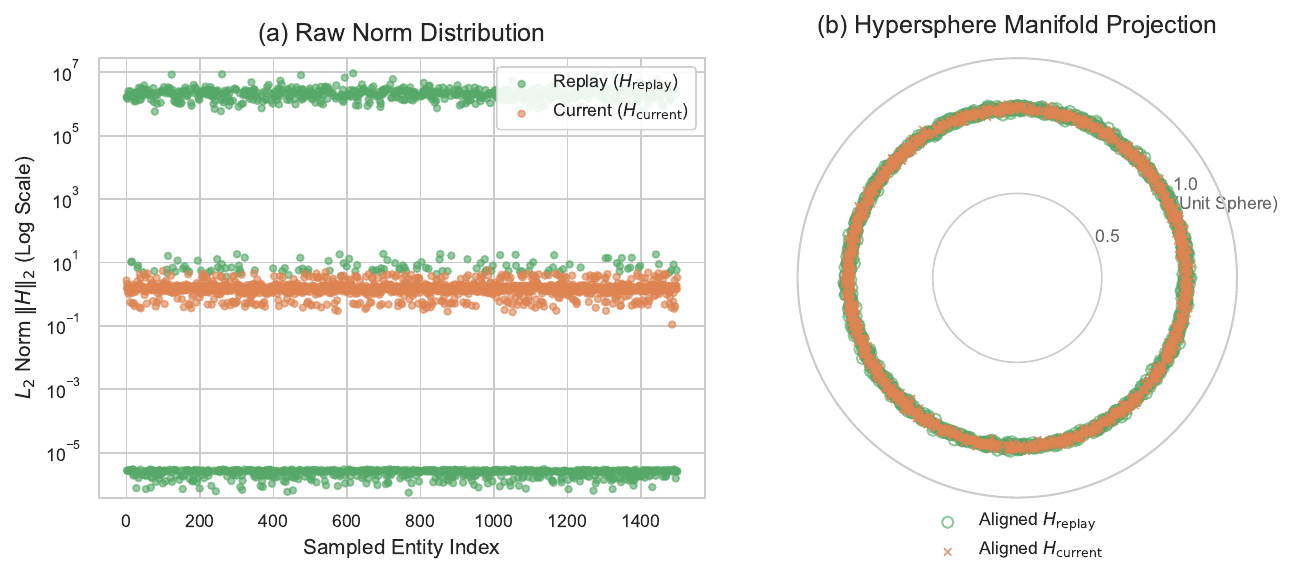}
\caption{Analysis of feature norms. (a) L2 norm distributions of raw current and replay representations. (b) Projection onto the unit hypersphere manifold via Manifold Alignment.}
\label{fig:norm_analysis}
\vspace{-6mm}
\end{figure}

To answer RQ4, we visualize the L2 norms of the hidden representations before and after Manifold Alignment. As depicted in Figure \ref{fig:norm_analysis}(a), the raw feature spaces exhibit noticeable \textit{norm domination}. Specifically, the historical replay representations ($H_{replay}$) exhibit a different magnitude distribution compared with the current representations ($H_{current}$), resulting in an imbalance during direct feature fusion. This magnitude discrepancy indicates that directly combining the raw representations may cause one representation source to dominate the fusion process. If these raw representations are directly fused, the magnitude-dominant representation may bias the optimization process and weaken the contribution of the other representation source.

As shown in Figure \ref{fig:norm_analysis}(b), Manifold Alignment projects these heterogeneous representations onto a uniform hypersphere manifold ($||H||_2 = 1$). This normalization reduces the scale discrepancy between historical and current representations, alleviating \textit{norm domination} before subsequent semantic routing.

\FloatBarrier
\subsection{Regularized Gating Analysis (RQ5)}

\begin{figure}[t]
\centering
\includegraphics[width=\linewidth]{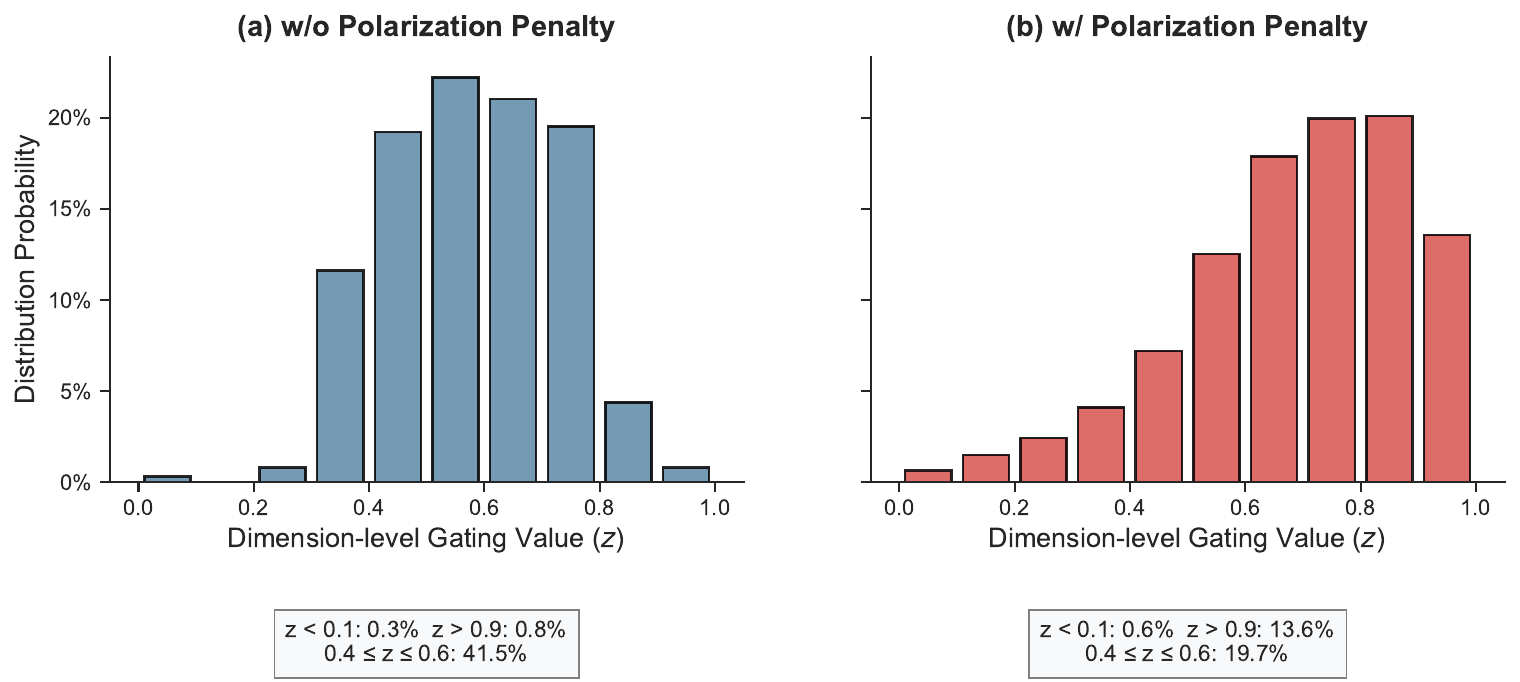}
\caption{Distribution of dimension-level gating values $z$ on ICEWS14s using RE-GCN equipped with MA-DAR under the diffusion replay setting. (a) Standard soft-gating without the polarization penalty. (b) Gating distribution with the polarization penalty, encouraging more decisive dimension-wise routing decisions.}
\label{fig:gating_analysis}
\vspace{-4mm}
\end{figure}

\begin{figure}[t]
\centering
\includegraphics[width=\linewidth]{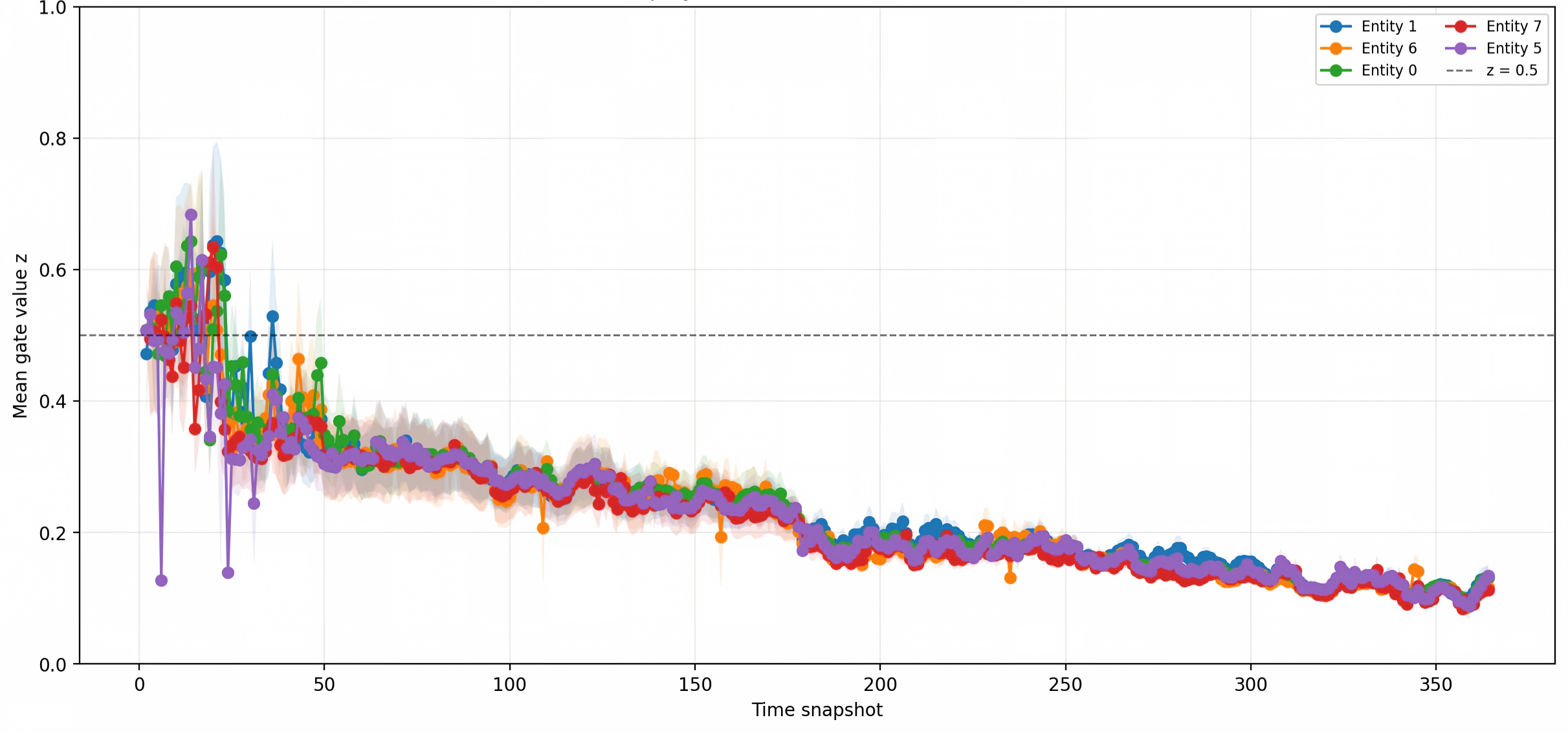}
\caption{Case study of temporal gating evolution for representative entities on ICEWS14s using RE-GCN equipped with MA-DAR under the diffusion replay setting.}
\label{fig:gate_case}
\vspace{-6mm}
\end{figure}

To address RQ5, we extract and visualize the test-phase distribution of dimension-level gating weights $z \in [0, 1]^d$ on ICEWS14s using RE-GCN equipped with MA-DAR under the diffusion replay setting. As illustrated in Figure \ref{fig:gating_analysis}(a), soft gating without $\mathcal{L}_{polar}$ regularization leaves a large proportion of gating values (41.5\%) in the ambiguous middle zone ($0.4 \le z \le 0.6$), which may lead to \textit{semantic blurring} during representation fusion.

With the proposed Polarization Penalty, the proportion of ambiguous intermediate values is reduced to 19.7\% (Figure \ref{fig:gating_analysis}(b)), encouraging gating values toward clearer binary-like decisions. This distribution indicates that MA-DAR produces more decisive dimension-wise routing patterns, reducing uncertain intermediate allocations during representation fusion. Unlike a non-differentiable hard threshold, the proposed penalty encourages decisive routing while preserving end-to-end gradient-based optimization.

Beyond the statistical distribution, we further provide a case study to illustrate how MA-DAR dynamically adjusts gating behaviors during temporal evolution. As shown in Figure \ref{fig:gate_case}, we track the gating trajectories of representative entities on ICEWS14s using the same RE-GCN-based MA-DAR configuration under diffusion replay. The trajectories show that gating values vary across timestamps rather than remaining fixed, indicating that MA-DAR dynamically adjusts the contributions of historical and current representations according to evolving temporal contexts.

\section{Conclusion}

In this paper, we introduced MA-DAR, a general plug-and-play routing for replay representation fusion in continual TKG reasoning. By decoupling historical representation integration from backbone temporal encoding, MA-DAR can be flexibly combined with diverse architectures, including RE-GCN, TiRGN, and LogCL. Through Manifold Alignment, Dynamic Gating, and Polarization Penalty, MA-DAR effectively resolves representation conflicts between historical and current features by alleviating \textit{norm domination} and \textit{semantic blurring}. Extensive experiments under the strict time-aware filter protocol demonstrate that MA-DAR consistently improves different backbone encoders and establishes new state-of-the-art performance across multiple benchmarks.

\bibliography{aaai2027}

\end{document}